\documentclass[journal]{IEEEtran}

\usepackage{amssymb}
\usepackage{amsmath}
\usepackage{makeidx}
\usepackage{multicol}
\usepackage[abs]{overpic}
\usepackage{tabularx}
\usepackage[numbers,sort&compress]{natbib}
\usepackage{amsfonts}
\usepackage{color}
\usepackage{array,color}
\usepackage{tabularx}
\usepackage{multirow}
\usepackage{graphics}
\usepackage{epsfig}
\usepackage{graphicx}
\usepackage{makeidx}
\usepackage{multicol}
\usepackage{amsthm}
\usepackage{bm}
\usepackage{booktabs}
\usepackage{ulem}
\usepackage{enumerate}
\usepackage{subfig}

\newtheorem{proposition}[theorem]{Proposition}

\begin{document}

\title{Individual Recognition in Schizophrenia using Deep Learning Methods with Random Forest and Voting Classifiers: Insights from Resting State EEG Streams}

\author{Lei Chu,
        Robert Qiu, Haichun Liu, Zenan Ling, Tianhong Zhang and Jijun Wang
\IEEEcompsocitemizethanks{\IEEEcompsocthanksitem Dr. Qiu is with the Department of Electrical and Computer Engineering\, Tennessee Technological University, Cookeville, TN 38505 USA. Dr. Qiu is also with Department of Electrical Engineering, Research Center for Big Data Engineering Technology, Shanghai Jiaotong University, Shanghai 200240, China. (e-mail: rqiu@tntech.edu). \protect
\IEEEcompsocthanksitem Lei Chu, Haichun Liu and Zenan Ling are with Department of Electrical Engineering, Research Center for Big Data Engineering Technology, Shanghai Jiaotong University, Shanghai 200240, China. (leochu, haichunliu, zenanlin, dugushixin@sjtu.edu.cn)}
\thanks{Dr.  Qiu's  work  is partially supported  by  N.S.F. of  China  No.61571296 and N.S.F.  of  US  Grant No.  CNS-1247778, No.  CNS-1619250. }}

\IEEEtitleabstractindextext{%
\begin{abstract}

Recently, there has been a growing interest in monitoring brain activity for individual recognition system. So far these works are mainly focussing on single channel data or fragment data collected by some advanced brain monitoring modalities. In this study we propose new individual recognition schemes based on spatio-temporal resting state Electroencephalography (EEG) data. Besides, instead of using features derived from artificially-designed procedures, modified deep learning architectures which aim to automatically extract an individual's unique features are developed to conduct classification. Our designed deep learning frameworks are proved of a small but consistent advantage of replacing the $softmax$ layer with Random Forest. Additionally, a voting layer is added at the top of designed neural networks in order to tackle the classification problem arisen from EEG streams. Lastly, various experiments are implemented to evaluate the performance of the designed deep learning architectures; Results indicate that the proposed EEG-based individual recognition scheme yields a high degree of classification accuracy: $81.6\%$ for characteristics in high risk (CHR) individuals, $96.7\%$ for clinically stable first episode patients with schizophrenia (FES) and $99.2\%$ for healthy controls (HC).

\end{abstract}

}

\maketitle

\IEEEpeerreviewmaketitle
\IEEEdisplaynontitleabstractindextext

\section{Introduction}
\label{Sec:1}


\IEEEPARstart{S}{chizophrenia} today is a chronic, frequently disabling mental disorder that affects
about one per cent of the world¡¯s population \cite{Insel2010Rethinking}; And it is widely perceived as one of the most severest mental disorder compromising multi-aspect of everyday quality of life \cite{Aswin2016Schizophrenia}. This predicament often continues in spite of pharmacological treatment of psychotic symptoms \cite{Gould2001Cognitive}. Accordingly, increasing attention is paid to the studies on individual recognition in schizophrenia (IRS) with the aim of surveillance, early detection or pre-diagnosis.

An typical IRS scheme consists of two phases: off-line training and online recognition \cite{Ula2011Heat, Choi2016Biometric}. During the off-line training phase, knowledgeable and unique individual characteristics are measured and recorded by some advanced brain monitoring modalities which includes EEG, electrical impedance tomography (EIT), Magnetoencephalography (MEG), Quantitative susceptibility mapping (MAP), electroneurogram (ENG), etc. These modalities promise to piece together different factors of the brain and provide new insights to help detect and treat diseases; So they are particularly appropriate for a disease as schizophrenia which impacts many aspects of the brain \cite{Hassanien2014Brain}. In this context, however, we use EEG as brain monitoring modality for the following considerations:

\begin{enumerate}
\item EEG provides a high spatio-temporal resolution data, a vivid reflection of dynamics of the brain \cite{Guger2001Rapid}.
\item EEG is the most inexpensive method of neuroimaging which plays a fundamental role for implementing deep learning methods \cite{Panayiotopoulos2010EEG}.
\item The EEG of a normal and healthy brain will differ from a brain with disease or functioning abnormally or in different healthy condition \cite{Knyazeva2001EEG}.
\item EEG shows small intra-personal differentiation and large inter-personal differentiation \cite{Guevara1995Inter}.
\item A number of brain diseases are feasible to diagnose, study and analyze by the EEG; The diseases includes Headache, Parkinson's disease, Schizophrenia, Attention Deficit Hyperactivity Disorder, etc. See monographs \cite{Panayiotopoulos2010EEG, Hassanien2014Brain} for the more detailed summary.
\end{enumerate}

Note that preprocessing for these raw data is also accomplished in this phase. The "brain fingerprinting" is then constructed for every kind of candidate. During the online recognition phase, various methods utilizing the recorded data and their extracted features can be applied to classify the candidates when the online individual characteristics were collected and refined.

\subsection{Related Works and Motivations}
\label{rwm}

EEG-based biometrics offer an exciting new form of human computer interface where a device can be controlled and provide available data for individual recognition analysis. Related works on EEG-based individual recognition analysis are summarized briefly as follows.

So far, manually-designed experimental protocols and EEG features that have been commonly utilized for the devising of EEG biometric systems aimed to recognize characteristics of spatially limited sets of brain regions.
Nakayama and Abe discuss the feasibility of using single-channel EEG waveforms for single-trial classification of viewed characters \cite{Nakayama2010Feasibility}. Berthomier and coauthors present an automatic analysis of single-channel frequency EEG measurements for validation in healthy individuals \cite{Berthomier2007Automatic}. The monograph gives a comprehensive study of classification of EEG signals using single channel independent component analysis, power spectrum, and linear discriminant analysis \cite{Tjandrasa2016Classification}. To take advantage of spatial information provided by multi-channel EEG and obtain higher classification accuracy, Krajca $et. al.$ achieve an automatic identification of significant graphoelements in multi-channel EEG recordings by adaptive segmentation and fuzzy clustering \cite{Krajca1991Automatic}. Prasad and coauthors realize a single-trial eeg classification using logistic regression based on ensemble synchronization; These works could classify each single trial of EEG as belonging to a patient with schizophrenia or a healthy control subject with 73\% accuracy \cite{Prasad2014Single}.
However, these works are based on features extracted from single channel or multi-channel EEG fragment and fail to obtain the accurate and robust classification result, thus are unfeasible for practices. Fortuately, recently advanced big data analysis based on streaming data \cite{Landhuis2017Neuroscience,Chu2016Voltage,qiu2015smart} could provide new means and ideas for the planning and design of classification scheme.

In this paper we conduct the IRS task based on resting state EEG. There are two reasons for that. First, evidence suggests that electrical activities¡¯ resting state organizes and coordinates neuronal functions \cite{K2011Resting}. Second, certain tasks cannot be performed by certain group of people, e.g., schizophrenia, Attention Deficit Disorder, or hyperactivity disorder \cite{Karatekin1998Working}.

The difficulty encountered in resting state EEG based IRS scheme is that resting state EEG \cite{Sponheim1994Resting} streams lack task-related feature, thus leading to a hard task to obtain the best and unique feature for an individual. Accordingly, there has been emerging a great need for the capability to extract features automatically. Kottaimalai and coauthors put forward EEG signal classification using Principal Component Analysis with Neural Network in Brain Computer Interface applications \cite{Kottaimalai2013EEG}. Li and Fan suggest a classification method to separate Schizophrenia and depression by EEG with artificial neural networks (ANN) \cite{Li2006Classification}. Ruffini et al. present EEG-driven classification for Prognosis of Neurodegeneration in At-Risk Patients by recurrent neural networks (RNN) \cite{Ruffini2016EEG}. ANN-based and RNN-based neural network structures require the non-vectorial inputs such as matrices to be converted into vectors which has been proved of problematic \cite{DBLP:journals/corr/GaoGW16, Schmidhuber2015Deep}. The vectorization of EEG streams would lose spatio-temporal information and give a very large solution space that demands very special treatments to the network parameters and high computational cost. As novel alternatives, convolution neural network (CNN) can help improve a learning system with three advantages£ºsparse interactions, parameter sharing and equivariant representations \cite{Lecun2015Deep}. Recently, Ma et al. conduct resting state EEG-based biometrics for individual identification using convolutional neural networks \cite{Ma2015Resting}; And their results indicate that the CNN-based joint-optimized EEG-based biometric system yields a high degree of accuracy of identification (88\%) which still can not reach the practical requirement. In summary, to obtain a higher classification accuracy, combining CNN-based network structure with spatiotemporal EEG analysis will be indispensable.

\subsection{Our Contributions}

Based on considerations above, we propose a new IRS scheme using advanced deep learning methods, aiming at automatically extracting features and performing classification. Our main contributions are summarized as follows.

\begin{enumerate}
\item Instead of utilizing short term EEG data which were found insufficient to provide required information for IRS analysis, we employ streaming EEG data collected by multi-channel scalp electrodes.
\item Three kinds of advanced deep learning methods were developed for IRS analysis.
\item The $softmax$ classifier which was widely used in classical deep learning methods is replaced by RF with aim of improving classification accuracy.
\item To tackle the classification problem with EEG data streams, a voting layer is developed at the top of the employed neural networks.
\item Various experiments are conducted to investigate the effectiveness and robustness of proposed IRS scheme.
\end{enumerate}

The remainder of this paper is structured as follows. Section \ref{Sec:2} firstly introduces the procedure of collection, preprocessing and mathematical representation for EED data streams; The proposed IRS scheme based on advanced deep learning methods is then developed in the second part of Section \ref{Sec:2}. In Section \ref{sec:case}, numerical case studies are provided to evaluate the performance of the proposed IRS scheme. Conclusion and acknowledgement of this research is given in Section \ref{sec:Conclusion} and Section \ref{sec:ack}, respectively.

\section{Materials and Methods}
\label{Sec:2}

\subsection{EEG Data Collection}
\label{data-collect}

The present work aimed to study IRS issue by assessing three types of subjects: characteristics in high risk (CHR) individuals, clinically stable first episode patients with schizophrenia (FES) and healthy controls (HC). 120 subjects (40 CHRs, 40 FESs and 40 HCs) were included; And all subjects to be investigated in this context were recruited from outpatients at Shanghai Mental Health Center. All subjects were free of mental retardation, neurological diseases, substance abuse or alcohol and any physical illness that may influence their cognitive function. The study protocol was approved by the Institutional Review Board of Shanghai Mental Health Center, and informed consents were obtained from all the subjects.

The experimental data were provided by Department of EEG Source Imaging in Shanghai Mental Health Center. So the data collection process is same with their previous work \cite{Sun2014Abnormal}. Participants were seated 1 m from the screen in a sound attenuated and electrically shielded chamber with dim illumination. EEGs were recorded from 64-channel scalp electrodes mounted in an elastic cap (BrainCap, Brain Products, Inc., Bavaria, Germany) including two pairs of vertical and horizontal electrooculography (EOG) electrodes. The electrode-scalp impedance was kept below 5 k¦¸ for each electrode.  Our analysis was performed on eye-open resting conditions, each single recording lasting over 300 seconds in time. Data recording was referenced to the tip of nose and sampled at $K=1000$ Hz.

\subsection{EEG Data Preprocessing and Mathematical Representation}
\label{data-prep}

\begin{figure}[!htp]
{
\includegraphics[width=0.475\textwidth]{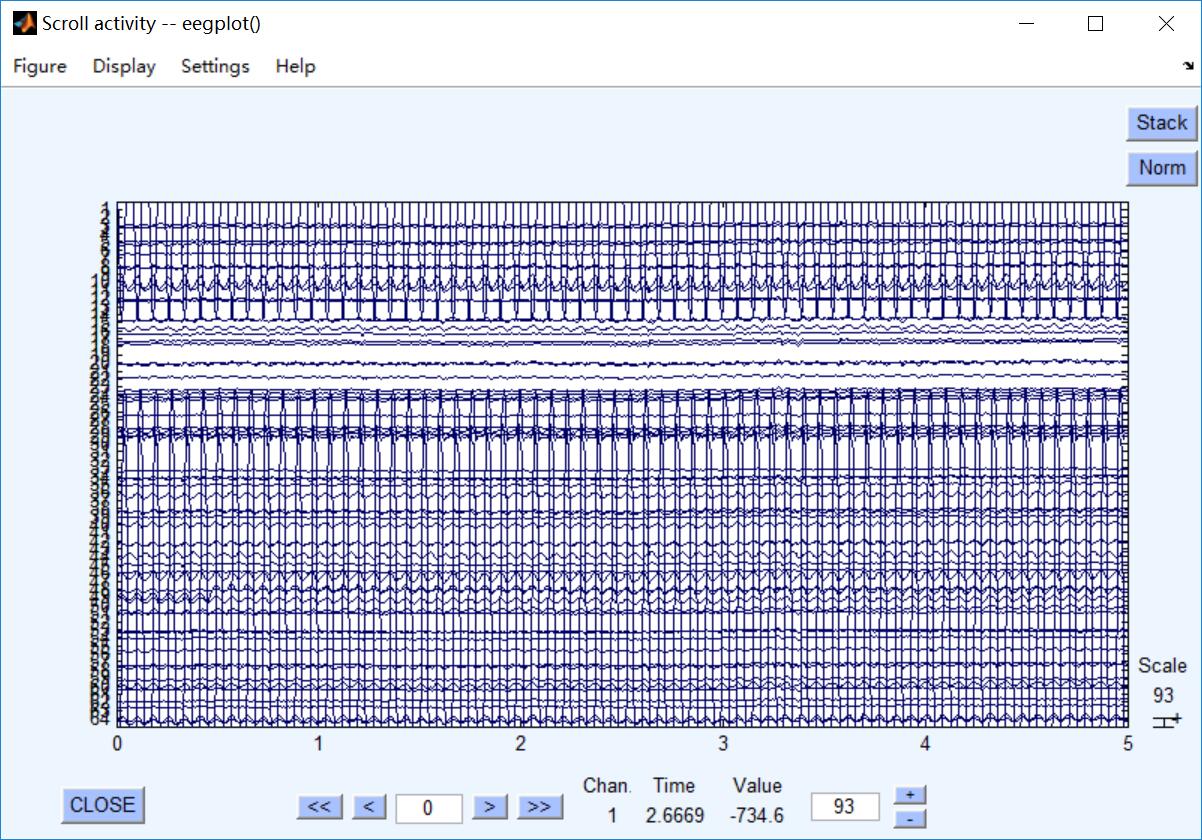}
}
\caption{{\label{fig_rd}} Raw eeg data.}
\end{figure}

\begin{figure}[!htp]
{
\includegraphics[width=0.475\textwidth]{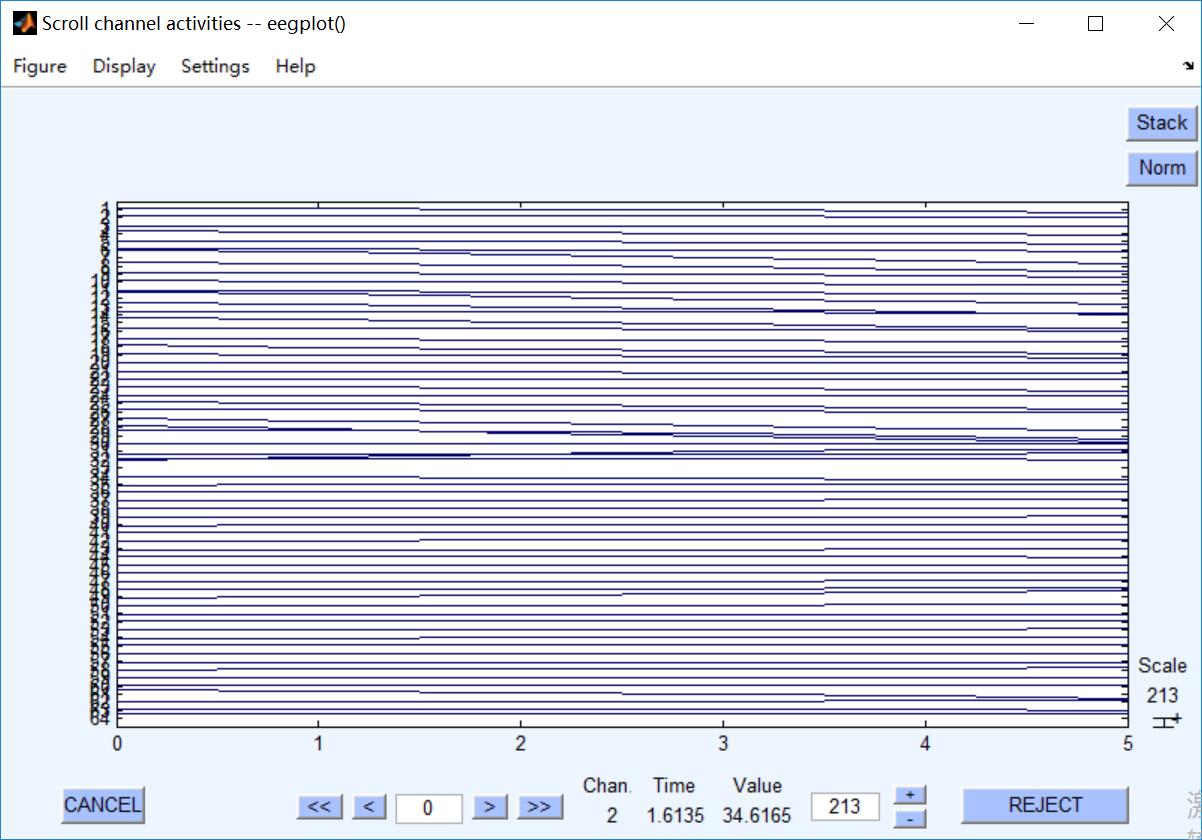}
}
\caption{{\label{fig_fd}} Filtered eeg data.}
\end{figure}

The brain vision analyzer (1.05, Brain Products, Inc.) was utilized for EEG preprocessing \cite{Lorist2009The}. Artifacts caused by vertical and horizontal eyes movements and blinks were removed off-line by an ocular correction algorithm [31]. All the artifact-reduced EEG data were referenced using the common average reference, band-pass filtered into 0.01¨C50 Hz using a zero phase-shift IIR filter (24 dB/Oct). See Fig. \ref{fig_rd} and Fig. \ref{fig_fd} for an illustration. After that, the broadband EEG signals containing artifacts were excluded using EEGLAB \cite{Brunner2013Eeglab}. After the preprocessing, in addition to preserve the complete signals, the EEG signals were also band-pass filtered into four classic frequency bands, i.e., $delta \ {\rm{(0.5 - 4)}}$ Hz, $theta \ {\rm{(4 - 8)}}$ Hz, $alpha \ {\rm{(8 - 13)}}$ Hz, $beta \ {\rm{(13 - 30)}}$ Hz and $gamma \ {\rm{(30 - 50)}}$ Hz bands, respectively, using least-squares FIR filters \cite{Vaidyanathan1987Nguyen}.

\begin{figure*}
\centering
\subfloat[FES subject]{ \label{FES-map}
\includegraphics[width=0.32\textwidth]{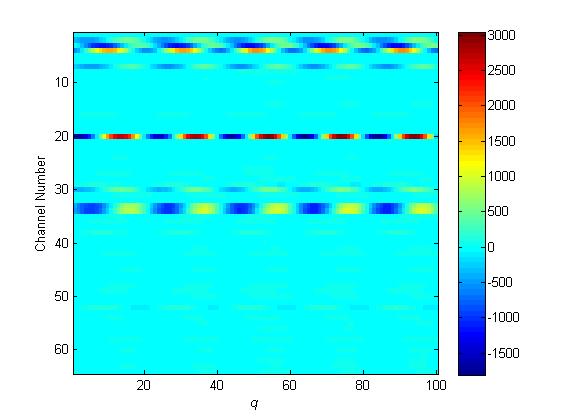}
}
\subfloat[CHR subject]{ \label{CHR-map}
\includegraphics[width=0.32\textwidth]{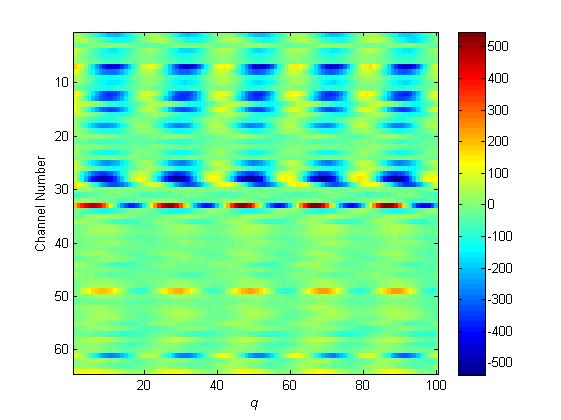}
}
\subfloat[HC subject]{ \label{HC-map}
\includegraphics[width=0.32\textwidth]{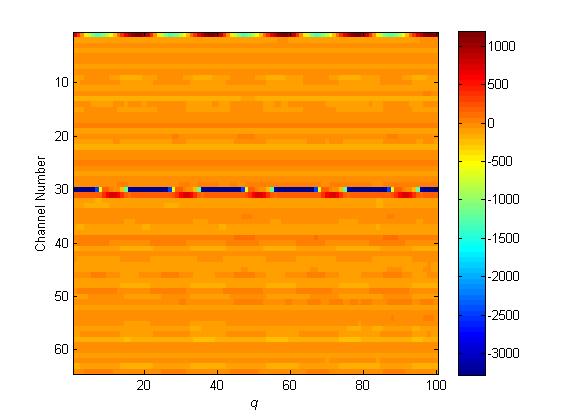}
}
\caption{{\label{fingers}} Examples of raw brain fingerprinting of three kinds of subjects (fragment size are: $p=64$ and $q=100$).}
\end{figure*}

In order to facilitate subsequent analysis, the mathematical representation of filtered EEG streams are described in the following. Let $m$ and $T$ denote the number of the available channel number of scalp electrodes and sampling time, respectively. To ensure the same length of collected EEG data in the following analysis, we have $T = 300s$ for all subjects. The total length of EEG data collected at every scalp electrode is $N = K*T = 300,000$. For $i$th type subject, a sliding window based data allocation scheme for the large EEG data matrix ${\bf{X}}^{(i)} \in \mathbb{R}^{m\times N}$ is presented as follows. Let $q$ be the sliding window size and $r = K/q$, then a sequence of matrix
\begin{equation}
\label{eq1}
\left\{ {\underbrace {{{\bf{X}}^{(i)}_{11}},{{\bf{X}}^{(i)}_{12}}, \cdots ,{{\bf{X}}^{(i)}_{1r}}}_{{\rm{1}}{\mathop{\rm st}\nolimits} \ {\rm{ sampling}}}, \cdots ,\underbrace {{{\bf{X}}^{(i)}_{T1}},{{\bf{X}}^{(i)}_{T2}}, \cdots ,{{\bf{X}}^{(i)}_{Tr}}}_{{\rm{Tth \ sampling}}}} \right\}
\end{equation}
is obtained to represent the collected EEG data streams. As shown in Fig. \ref{fingers}, these fragments ${\bf{X}}_{jk}^{(i)}$ are considered as raw Brain fingerprinting of all subjects.

\subsection{Proposed IRS Scheme based on Advanced Deep Learning Methods}
\label{DL}

\subsubsection{Classical Deep Learning Structures}

\begin{figure}[!htp]
{
\includegraphics[width=0.475\textwidth]{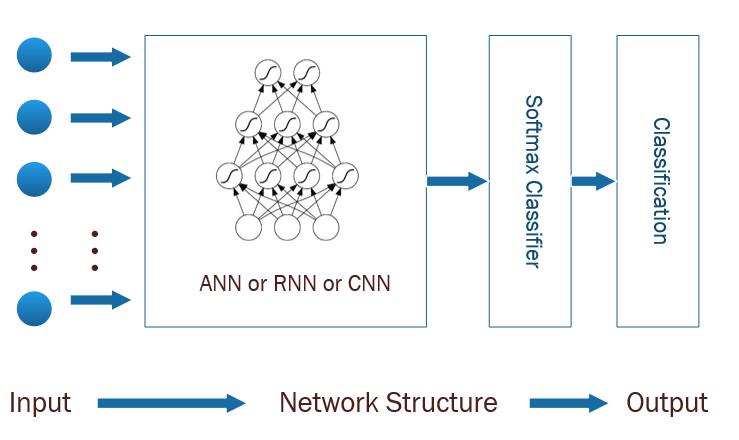}
}
\caption{{\label{bdls}} The basic deep learning structure.}
\end{figure}

As introduced in the Section \ref{rwm}, we have introduced previous biometrics that embrace classical deep learning methods, such as ANN, RNN, CNN and their modified versions. Here, some technical details are discussed in order to get better understanding on how to apply them. Input, a kind of network and $softmax$ classifier contribute to the basic elements of a deep learning structure (See Fig. \ref{bdls}). The implementation of classical deep learning methods for IRS problem is discussed in the following.

As introduced in the Section \ref{data-prep}, the collected EEG streams recorded for off-line analysis are represented by data fragments ${{\bf{X}}^{(i)}_{jk}}$, where $i=1,2,3$, $j=1,2, \cdots ,m$, $j=1,2, \cdots ,N$. These fragments were utilized to train the neural networks (ANN, CNN and RNN) after local normalization scheme represented by
\begin{equation}
{\bf{X}}_{jk}^{(i)}{\rm{ = }}{\bf{X}}_{jk}^{(i)}{\rm{/}}\max \left( {{\bf{X}}_{jk}^{(i)}} \right)
\end{equation}

For IRS classification problems based on deep learning methods, it is standard to use $softmax$ classifier at the top. Let the subjects be a finite space $\left\{ {{p_1},{p_2},{p_3}} \right\}$ with a finite observation space $\left\{ {{{\bf{X}}^{(1)}_{11}}, {{\bf{X}}^{(1)}_{12}}, \cdots {{\bf{X}}^{(3)}_{mN}},} \right\}$. Let $\Pr \left( {{{{\bf{X}}^{(i)}_{jk}}}|{p_i}} \right)$ be the learned model of the conditional probability of seeing observation data ${{\bf{X}}^{(i)}_{jk}}$ with people ${p_i}$. Let $\bf{b}$ and $\bf{W}$ be the activation of the penultimate layer nodes and the weight connecting the penultimate layer to a classifier layer, respectively. The total input into the classifier layer, denoted by $\bf{a}$, is
\begin{equation}
{a_l} = \sum\nolimits_h {{w_{hl}}{b_h}}.
\end{equation}
Given the $softmax$ classifier, we have
\begin{equation}
\Pr \left( {{\bf{X}}_{jk}^{(i)}|{p_i}} \right) = \frac{{\exp \left( {{a_h}} \right)}}{{\sum\nolimits_s {\exp \left( {{a_s}} \right)} }}
\end{equation}
The predicted class $i$ for the single fragment ${{\bf{X}}^{(i)}_{jk}}$ would be
\begin{equation}
{{\hat p}_i} = \mathop {\arg \max }\limits_i \Pr \left( {{\bf{X}}_{jk}^{(i)}|{p_i}} \right) = \mathop {\arg \max }\limits_i {a_i}
\end{equation}

\subsubsection{Deep Learning Methods using RF Classifier}

Most deep learning methods for classification utilizing convolutional and full-connected layers have used $softmax$ classifier to learn the small size parameters. There are exceptions, significantly in works \cite{Weston2012Deep, Weston2008Lecture}, supervised embedding with nonlinear NCA \cite{Taylor2010Pose}, semi-supervised deep embedding \cite{Weston2008Deep} and deep learning using linear support vector machines \cite{Gentile2001A}. In this paper, we replace the $softmax$ with RF for classification. RF has been studied extensively in the fields of nonparametric statistics \cite{Genuer2012Variance} and continue to be very popular because of its simplicity and because it is very successful for many practical problems \cite{M2005Random, D2006Gene}. Here we firstly summarize the basic principle of RF as follows.

Let $\bf{a}$ and $\bf{y}$ be the data features and the corresponding labels. RF is built from a training set $\left\{ {\left( {{a_h},{y_h}} \right)} \right\}_{h = 1}^n$ that make predictions ${{\bf{\hat y}}}$ for new points ${a'}$ by looking at the neighborhood of the point, formalized by a weight function ${\bf{H}}_l$ in $l$th tree:
\begin{equation}
{{\bf{\hat y}}_l} = \sum\limits_{h = 1}^n {{\bf{H}}_l\left( {{a_h},a'} \right){y_h}}.
\end{equation}
Here, ${{\bf{H}}_l\left( {{a_h},a'} \right)}$ is the non-negative weight of the $h$th training point relative to the new point ${a'}$ and $n$ is the number of nodes in the penultimate layer of the neural network in this work. For any ${a'}$, the weights sum to one. Since a forest averages the predictions of a group of $M$ trees with individual weight functions ${\bf{H}}_l$, its predictions are
\begin{equation}
{\bf{\hat y}} = \frac{1}{M}\sum\limits_{l = 1}^M {\sum\limits_{h = 1}^n {{{\bf{H}}_l}\left( {{a_h},a'} \right){y_h}} },
\end{equation}
then the predicted class is
\begin{equation}
{{\hat p}_i} = \arg \max {\bf{\hat y}}.
\end{equation}
For more technical details about RF, interested readers are referred to the distinguished works by Breiman \cite{Breiman2001Random, Breiman2004Consistency}.

Another advanced classifier, a linear multi-class support vector machine (mSVM) which has been proved of higher classification accuracy in \cite{Arunkumar2015Multi, Tang2015Deep}, is also adopted for the purposes of comparison. We verified the effectiveness of the proposed scheme on the well-known $mnist$ dataset: a nine-layer CNN achieved a $1.93\%$ test error with RF classifier, $1.95\%$ with mSVM classifier and $2.27\%$ with $softmax$ classifier.

\subsubsection{Streaming EEG Classification with a Voting Scheme}

It is worth noting that the above modified deep learning methods are suitable to classify subjects with single fragment ${{\bf{X}}^{(i)}_{jk}}$. Let $\pi _{jk}^{(i)} = \Pr \left( {{\bf{X}}_{jk}^{(i)}|{p_i}} \right)$. To handle the scenario that the subjects are with EEG streams ${\bf{X}}^{(i)}$, we develop a voting layer whose decision rule is denoted by
\begin{equation}
{{\tilde p}_i} = \Pr \left( {{{\bf{X}}^{(i)}}|{p_i}} \right) = \arg \max \sum\nolimits_{j,k} {\Pr \left( {{\bf{X}}_{jk}^{(i)}|{p_i}} \right)} /Q,
\end{equation}
where $Q = N/r$.

Moreover, in this paper, we also adopt the some other recently developed techniques to improve the performance of deep learning methods employed in the IRS analysis. Specially, we use exponential linear unit (ELU) proposed in \cite{Shah2016Deep} to accelerate the learning speed in deep neural networks. The $max \ pooling$ technique \cite{Krizhevsky2012ImageNet} is utilized to prevent substantial overfitting problem.

Lastly, for the readers' convenience, we give a brief summary of the modified deep learning methods employed in this context in the following two tables (Tab. \ref{table:SNN} and Tab. \ref{table:DL}).

\begin{table}[ht]
\caption{{\label{table:SNN}} THE STRUCTURES OF NEURAL NETWORKS.}
\begin{tabularx}{0.48\textwidth}{|p{.8cm}<{\centering} | p{2.25cm}<{\centering}  | p{2cm}<{\centering} | p{1.9cm}<{\centering}|} 
\toprule
\multirow{1}{*}{}
& \multicolumn{3}{c}{$Methods$}  \\
 \cmidrule(lr){2-4}      
\multirow{1}{*}{$Layers$}
& \multicolumn{1}{c}{CNN} & \multicolumn{1}{c}{ANN} & \multicolumn{1}{c}{RNN}
\\
\midrule
1st     & Input & Input & Input  \\ \hline
2nd    & 2$\times$(Conv. + ELU) (kernel:$3\times3$, stride:2)  & Vectorization & Vectorization  \\ \hline
3rd    & Max Pooling (kernel:$2\times2$); Dropout(Rate:0.25)    & Dense(512) & recurrent layer (hidden units = 100)  \\ \hline
4th   & 2$\times$(Conv. + ELU) (kernel:$3\times3$)   & Activation('relu')  & Dense(3) \\ \hline
5th   & Max Pooling (kernel:$2\times2$, stride:2); Dropout(Rate:0.25)     & Dropout(Rate:0.25)  & $softmax$ or mSVM or RF       \\ \hline
6th    & 2$\times$(Conv. + ELU) (kernel:$3\times3$)     &  Dense(512)   & Output(Predict)   \\ \hline
7th   & Max Pooling (kernel:$2\times2$, stride:2); Dropout(Rate:0.25) & Activation('relu') & - \\ \hline
8th   & Dense(128) + ELU & Dropout(Rate:0.25)  & -   \\ \hline
9th  & Dropout(Rate:0.5)     & Dense(512) & -     \\ \hline
10th  & Dense(128) + ELU     & Activation('relu') & -    \\ \hline
11th   & Dropout(Rate:0.5)    &  Dropout(Rate:0.25)    & -  \\ \hline
12th  & Dense(3) + ELU    &  Dense(3)  & -    \\ \hline
13th  & Dropout(Rate:0.5)      & Dropout(Rate:0.5)    & -     \\ \hline
14th & $softmax$ or mSVM or RF    &  $softmax$ or mSVM or RF  &   -   \\ \hline
15th & Output(Predict)     & Output(Predict) &   -  \\ \hline
\bottomrule
\end{tabularx}
\end{table}

\begin{table}[ht]
\begin{center}
\caption{{\label{table:DL}} MODIFIED DEEP LEARNING METHODS}
\begin{tabular}{p{3cm}|p{4.5cm}}
\toprule
Deep Learning Methods & Explanation  \\
\midrule
ANNV   &  Classical ANN with $softmax$ classifier and a voting layer  \\
RNNV   &  Classical RNN with $softmax$ classifier and a voting layer  \\
CNNV   &  Classical CNN with $softmax$ classifier and a voting layer  \\
ANNV+mSVM   & Modified ANN using mSVM classifier and a voting layer  \\
RNN+mSVM   &  Modified ANN using mSVM classifier and a voting layer  \\
CNN+mSVM   &  Modified ANN using mSVM classifier and a voting layer  \\
ANN+RF   &  Modified ANN utilizing RF classifier and a voting layer  \\
RNN+RF   &  Modified RNN utilizing RF classifier and a voting layer  \\
CNN+RF   &  Modified CNN utilizing RF classifier and a voting layer  \\
\bottomrule
\end{tabular}
\end{center}
\end{table}

\section{Case Studies and Discussion}
\label{sec:case}

In this section, various experiments are developed to evaluate the performance of the proposed IRS schemes. We use the cross validation method to evaluate the performance of the proposed IRS scheme. Our results are the averages of 1000 independent run on GeForce GTX 750.

\subsection{The Accuracy of Time-domain and Frequency-domain EEG Data Streams}


%
%

Time-domain (as introduced in Section. \ref{data-prep}) and Frequency-domain (Amplitude of Fourier transform) EEG Data Streams are utilized firstly to perform and report accuracy assessments of proposed IRS schemes. The window size are set as $q=100$. The test data size are kept same with the training data for every subject. Tab. \ref{table:catf} shows that the proposed
CNNV+RF has the best classification accuracy against other methods.

\begin{table}[ht]
\caption{{\label{table:catf}} Classification Accuracy of Time-domain and Frequency-domain EEG Data Streams}
\begin{tabularx}{0.47\textwidth}{lccllll p{0.6cm}p{0.6cm}p{0.6cm}p{0.6cm}p{0.6cm}p{0.6cm}p{0.6cm}}
\toprule
\multirow{1}{*}{}
& \multicolumn{3}{c}{$Time-domain$}  & \multicolumn{3}{c}{$Frequency-domain$}  \\
 \cmidrule(lr){2-4}  \cmidrule(lr){5-7}      
\multirow{1}{*}{$Methods$}
& \multicolumn{1}{c}{FES} & \multicolumn{1}{c}{HC}
& \multicolumn{1}{c}{CHR} & \multicolumn{1}{c}{FES}
& \multicolumn{1}{c}{HC} & \multicolumn{1}{c}{CHR}
\\
\midrule
  ANNV        & 0.809    & 0.831    & 0.622               & 0.807    & 0.824     & 0.631  \\
 RNNV        & 0.742    & 0.803    & 0.594               & 0.731    & 0.792     & 0.588  \\
CNNV        & 0.923    & 0.952    & 0.755               & 0.915    & 0.949     & 0.749  \\
  ANNV+mSVM   & 0.811    & 0.841    & 0.643               & 0.804    & 0.929     & 0.639  \\
 RNNV+mSVM   & 0.759    & 0.826    & 0.602               & 0.744    & 0.807         & 0.589  \\
CNNV+mSVM   & 0.946    & 0.983    & 0.790               & 0.937    & 0.985     & 0.766  \\
  ANNV+RF     & 0.827    & 0.846    & 0.657               & 0.813    & 0.839     & 0.655  \\
 RNNV+RF     & 0.766       & 0.839    & 0.613            & 0.793    & 0.816     & 0.649  \\
CNNV+RF     & \bf{0.967}  & \bf{0.992} & \bf{0.816}     & 0.955    & 0.981     & 0.799  \\
\bottomrule
\end{tabularx}
\end{table}

\section{Conclusion}
\label{sec:Conclusion}

In conclusion, we have shown that CNNV-RF performs better than $softmax$ and CNNV-mSVM on a well-known dataset ($mnist$) and resting state EEG streams used in this paper. Switching from $softmax$ or mSVM to RF is incredibly simple and appears ro be helpful for classification problems. The experimental results show that the classification performance would be improved as the size of training and data database becomes larger. In the future, the proposed biometrics system should be tested on a larger group and more classes of subjects, providing further identification of accuracy, robustness and applicability of the system. The experiments also suggest that our results can be improved simply by waiting for faster GPUs.

\section{Acknowledgements}
\label{sec:ack}

We are appreciated for department of EEG source imaging leaded by professor Jijun Wang and professor Chunbo Li from SHJC for data providing and discussion. We also grateful to Dr. Tianhong Zhang from SHJC for his expert collaboration on data analysis.


\bibliographystyle{IEEEtran}
\bibliography{LeoChu}

\end{document}